\documentclass{article}

\usepackage{microtype}
\usepackage{graphicx}
\usepackage{subfigure}
\usepackage{booktabs} 
\usepackage{rotating}
\usepackage{amsfonts}

\usepackage{hyperref}

\usepackage[accepted]{icml2020}

\icmltitlerunning{Deep Semi-Supervised Embedded Clustering}

\begin{document}

\twocolumn[
\icmltitle{Deep Semi-Supervised Embedded Clustering (DSEC)\\ for Stratification of Heart Failure Patients}




\begin{icmlauthorlist}
\icmlauthor{Oliver Carr}{sh}
\icmlauthor{Stojan Jovanovic}{sh}
\icmlauthor{Luca Albergante}{sh}
\icmlauthor{Fernando Andreotti}{sh}
\icmlauthor{Robert D\"urichen}{sh}
\icmlauthor{Nadia Lipunova}{sh}
\icmlauthor{Janie Baxter}{sh}
\icmlauthor{Rabia Khan}{sh}
\icmlauthor{Benjamin Irving}{sh}
\end{icmlauthorlist}

\icmlaffiliation{sh}{Sensyne Health plc, Oxford, United Kingdom}

\icmlcorrespondingauthor{Benjamin Irving}{ben.irving@sensynehealth.com}

\icmlkeywords{Machine Learning, Clustering, Autoencoders, Embedding, Healthcare, Semi-supervised}

\vskip 0.3in
]



\printAffiliationsAndNotice{} 

\begin{abstract}
Determining phenotypes of diseases can have considerable benefits for in-hospital patient care and to drug development. The structure of high dimensional data sets such as electronic health records are often represented through an embedding of the data, with clustering methods used to group data of similar structure. If subgroups are known to exist within data, supervised methods may be used to influence the clusters discovered. We propose to extend deep embedded clustering to a semi-supervised deep embedded clustering algorithm to stratify subgroups through known labels in the data. In this work we apply deep semi-supervised embedded clustering to determine data-driven patient subgroups of heart failure from the electronic health records of 4,487 heart failure and control patients. We find clinically relevant clusters from an embedded space derived from heterogeneous data. The proposed algorithm can potentially find new undiagnosed subgroups of patients that have different outcomes, and, therefore, lead to improved treatments.

\end{abstract}

\section{Introduction}
\label{introduction}

Patient populations, such as heart failure, are often heterogeneous in presentation and responses to therapy. Heart failure is a complex disease typically classed into two categories by clinicians, one with reduced ejection fraction (HFrEF) or one with preserved ejection fraction (HFpEF) \cite{Inamdar2016}. Subgroups of heart failure patients can also be defined by additional measures that have been known to reflect poor outcome (e.g. serum urea, serum creatinine) or co-morbidities (e.g. diabetes) \cite{Inamdar2016}. Better characterisation of subgroups may allow for adjusted treatments in the different patient cohorts.

Embeddings, or low dimension representations of data are used to extract the most relevant features which describe the structure of the data. Low dimensional representation approaches are often based on natural language processing methods \cite{Zhang2018, Zhu2016} or autoencoders to produce latent space representations \cite{Wei2018}. Both Denaxas et al. \yrcite{Denaxas2018} and Choi et al. \yrcite{Choi2016} have investigated embeddings using heart failure patient data.

Electronic health record (EHR) data is heterogeneous, varies longitudinally, is sparse, and is not well suited to the application of traditional clustering analysis approaches without careful pre-processing \cite{Beaulieu-Jones2018, Donders2006}. Clustering approaches can be applied directly to the raw features or to an embedded representation of the input. Principal component analysis (PCA) \cite{Wold1987} is a common approach to finding a lower dimensional embedding but unsupervised or semi-supervised deep architectures such as autoencoders allow more complex representations of the data to be learnt \cite{Shickel2018,Beaulieu-Jones2016}. The use of continuous measures from EHRs have been used for patient subgrouping \cite{Liu2019,Choi2019}. These measures have been used in combination with medical codes, medication information, procedural codes, or text \cite{Li2019,Miotto2016, Lasko2013, Beaulieu-Jones2016}.

In this study we explore an extension of deep embedding clustering approaches to EHR patient cohorts in both a unsupervised and semi-supervised fashion. This work builds on previous approaches such as Deep Patient \cite{Miotto2016} and Beaulieu-Jones et al. \yrcite{Beaulieu-Jones2016} which use standard autoencoders. Previous work has also focused on semi-supervised clustering, such as Enguehard et al. \yrcite{Enguehard2019} who use convolutional neural networks to cluster partially labelled images and Ren et al. \yrcite{Ren2019} who use pairwise constraints to cluster data, whereas in this work we focus on transfer learning of a network. The key contributions of this paper are i) the application of DEC to patient records, ii) the extension of DEC as a novel semi-supervised approach that allows the handling of heterogeneous input measures and, iii) we show that clinically relevant subgroups within a heart failure cohort can be determined using data driven approaches.

\section{Methods}
\subsection{Deep Embedded Clustering (DEC)}

DEC \cite{Xie2015} is a powerful approach that combines an autoencoder with a clustering loss to learn a representation of the data that aims to also produce separable clusters to improve the analysis of the embedded space.

The DEC method transforms the \textit{data space} $X^n$ using a non-linear mapping $f_\theta : X^n \to Z^m$, where $n=n_{feat}$ is the number of features used and $Z^m$ is the \textit{embedded space}, and is a lower dimensional space than $X^n$ ($m<n)$. $\theta$ is a set of learnable parameters and $f_\theta$ is parametrized as a deep neural network.

Initially, a multi-layer deep autoencoder is implemented as a series of stacked de-noising autoencoders, with a bottleneck layer acting as the embedded space. Rectified linear unit (ReLU) activation functions \cite{Nair2010} are used at each layer except the first and last layers. The multi-layer deep autoencoder is then trained to optimize a reconstruction loss, by minimizing the least squares error between the input and output of the autoencoder. Once the autoencoder has been pre-trained, the \textit{decoder} is cut off leaving the \textit{encoder} to transform the input, $X^n$ to the embedded space, $Z^m$. 

Further, a clustering step is then added to the network after the embedded space which uses k-means clustering in the embedded space, $Z^m$. For this purpose, the network is fine tuned using the Kullback-Leibler (KL) divergence as loss function for the clustering step. An iterative approach is used to assign points in the embedded space, $Z^m$ to the $k$ cluster centroids, $\{\mu_j\}_{j=1}^k$, and to update the non-linear mapping, $f_\theta$, of $X^n$ to $Z^m$. This step is repeated until a convergence criteria is met, full details of method are shown in \cite{Xie2015}.

\subsection{Deep Semi-Supervised Embedded Clustering (DSEC)}

While autoencoders are an effective approach for learning a lower dimensional embedding, a key challenge of unsupervised approaches to heterogeneous data is that, while the embedding may be representative of the inputs, the representation might not accurately reflect the features of interest for patient or disease stratification. This can be partially resolved by applying a supervised model to the embedding \cite{Beaulieu-Jones2016} or with transfer learning of a pre-trained network \cite{Han2019}. We propose, modifying the latent representation on known patient subgroups by transfer learning of the encoder and fine-tuning of layers. This adapts the embedding to the problem of interest.

We propose to train the DSEC model in three sequential steps: training an autoencoder, updating the weights of the encoder with a classification task, and updating all the layers of the encoder with a clustering loss.

In our approach, a de-noising autoencoder is used to determine the embedded space, which partially corrupts the input, $X^n$, before reconstruction of the original data \cite{Vincent2010}. The input data $X^n$, $X\in \mathbb{R}^k$, is corrupted to $\tilde{X}^n$ through a stochastic mapping $\tilde{X} \sim q_D(\tilde{X}|X)$ by the addition of a Gaussian noise layer to the input of the de-noising autoencoder. The reconstruction error is measured by the MAE loss.

The non-linear mapping, or encoder, is represented by two fully connected dense layers, the first with 1000 nodes and the second with 500 nodes, and the embedded space having three nodes. The number of nodes were chosen by adapting the DEC architecture \cite{Xie2015}. Each layer uses ReLU activation functions. The Adam optimizer (learning rate $=0.01$, $\beta_1=0.9$, $\beta_2=0.999$) is used to optimise the weights of the network \cite{Kingma2015}.

Once the de-noising autoencoder has been pre-trained, the decoder is removed and a fully connected classification layer with softmax activation is added to the encoder. The de-noising corruption is removed from the architecture and the weights of the first dense layer are fixed. Known labels for each observation are used to update the weights of the final dense layer based on the classification task. Transfer learning updates the weights of the final dense layer and the embedding layer to minimize the binary cross-entropy loss function. This results in an updated mapping of the input $X$ to a new embedded space, $Z^\prime$.

A clustering loss (Kullback-Leibler divergence) is then used further update the entire encoder and latent space, updating the non-linear mapping to $f_\theta^{\prime\prime} : X \to Z^{\prime\prime}$. The optimization of the encoder and cluster centers is performed as in \cite{Xie2015}. The cluster centers, $\{\mu_j\}$, and the encoder parameters, $\theta$, are jointly optimized using the Adam optimizer (learning rate $=0.01$, $\beta_1=0.9$, $\beta_2=0.999$).

The number of epochs used for training were determined from the loss of the validation set in order to avoid overfitting to the training set. For DEC, the autoencoder was trained for 50 epochs and the clustering step for 200 epochs. For DSEC, the autoencoder was trained for 50 epochs, the semi-supervised transfer learning for 10 epochs, and the clustering for 200 epochs. The models were then trained on the entire training set before being applied to the test set. 

\subsection{Analysis of the Embedded Space}

\begin{figure*}[t]
\vskip 0.2in
\begin{center}
\centerline{\includegraphics[width=0.75\textwidth]{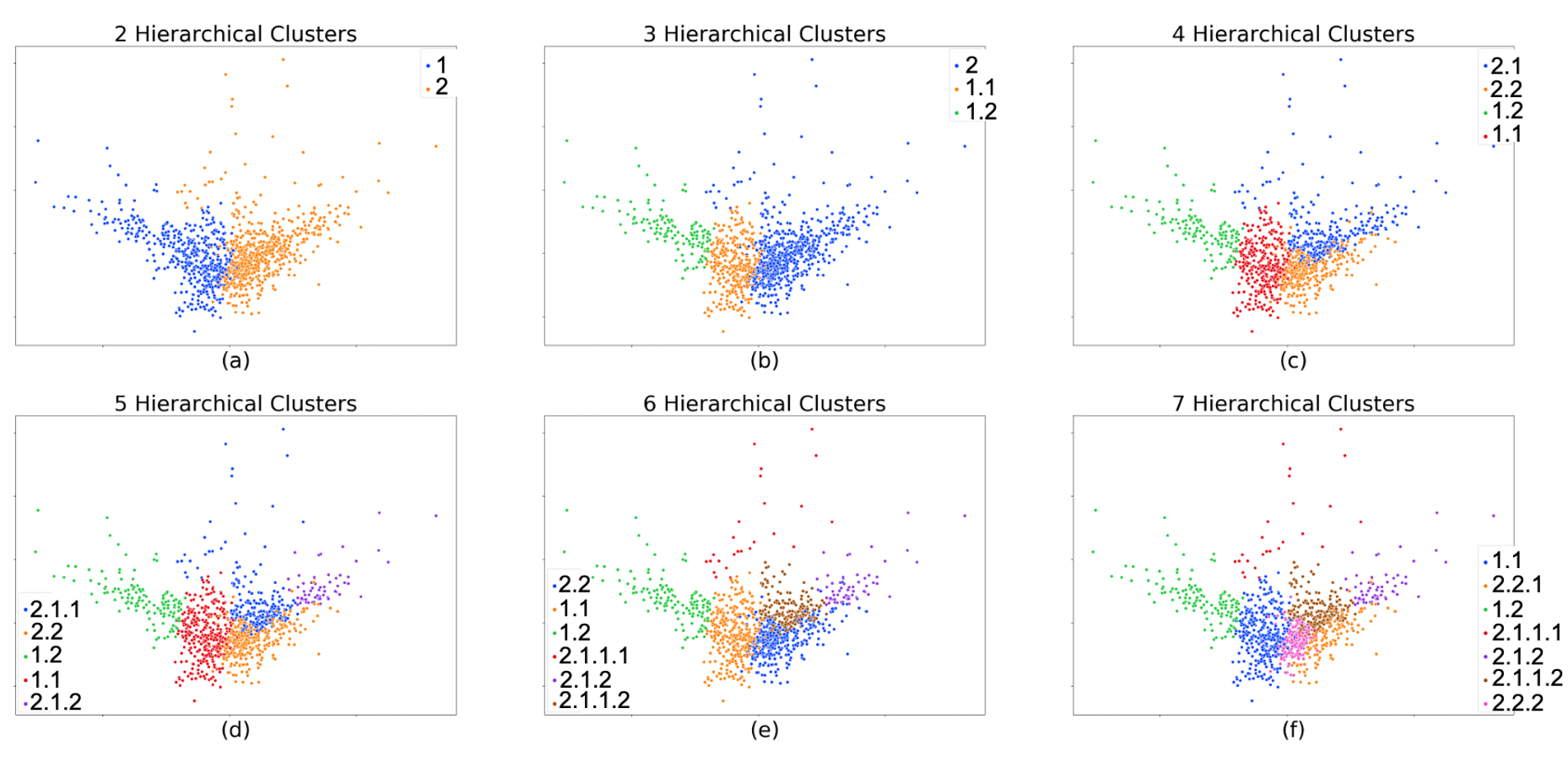}}
\caption{Hierarchical clustering of vital signs and laboratory measures shown as a PCA projection of the three-dimensional embedded space from DSEC.}
\label{hierarchical}
\end{center}
\vskip -0.2in
\end{figure*}

\begin{figure}[t]
\vskip 0.2in
\begin{center}
\centerline{\includegraphics[width=0.9\columnwidth]{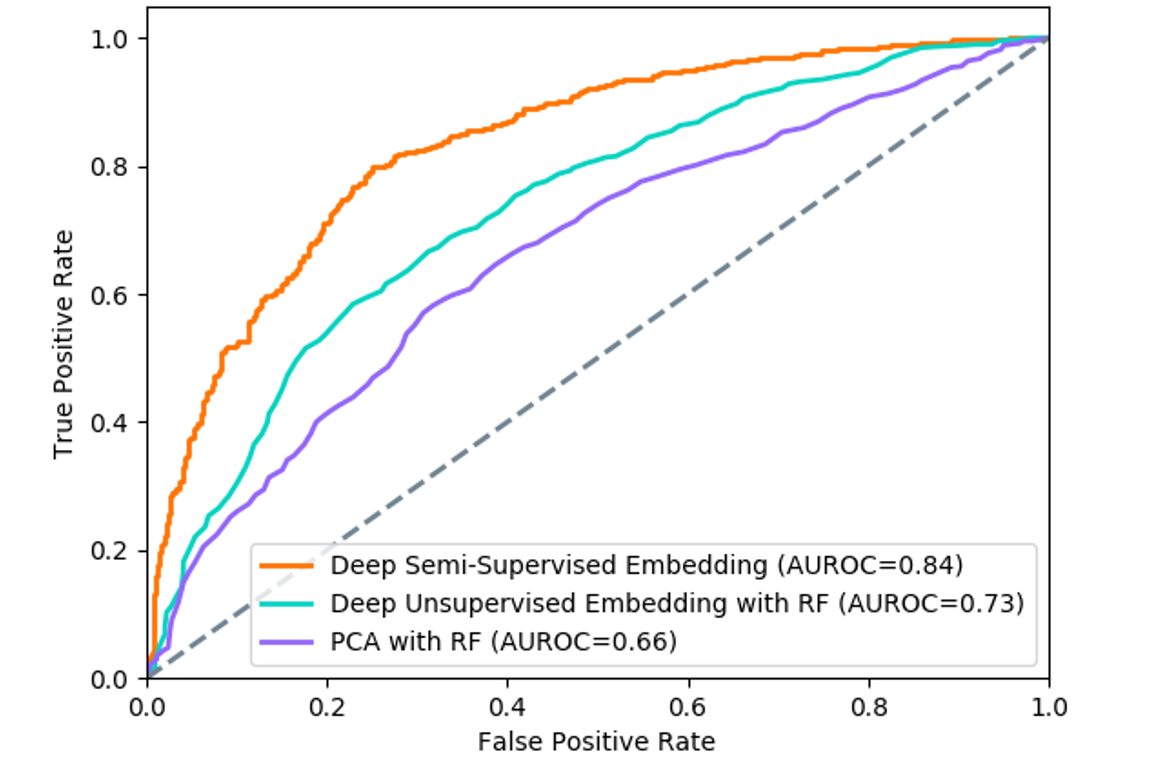}}
\caption{Receiver operating characteristic curves for classification of heart failure and control patients. Curves are shown for PCA and random forest, DEC and random forest, and DSEC.}
\label{roc}
\end{center}
\vskip -0.2in
\end{figure}

After a low dimensional patient representation is created using DSEC, we define the patient subgroups using standard clustering approaches on the embedded space. Agglomerative clustering is a type of hierarchical clustering method in which all observations initially start as individual clusters before pairs of clusters are successively merged into a new cluster \cite{Rokach2005}, with the Ward criteria used as the linkage function \cite{Ward1963}.

Once the subgroups are found we compared the dominant ICD-10 diagnosis codes in each subgroup using enrichment analysis. Enrichment analysis is performed using the Fisher exact test \cite{Fisher1922} for pairwise comparisons between ICD-10 codes within clusters. For each ICD-10 code the log odds ratio and corresponding p-value are found. The p-values are corrected for multiple testing using Bonferroni correction, with statistically significant odds ratios indicating the ICD-10 code is enriched in one of the clusters. For hierarchical clustering, enrichment is performed pairwise between the two clusters which are combined in each of the agglomeration steps.

\section{Data}

De-identified patient health record data was obtained from a large UK general trust hospital. These patients underwent digital monitoring of bedside vital sign measurements \cite{Wong2017}. From this dataset, patients with a primary diagnosis of heart failure (ICD-10 code I50*) were selected. Each patient has a number of admissions that may occur before, on or after a heart failure diagnosis. This resulted in 2,791 patients with 27,143 admissions. We used propensity matching (based on logistic regression and nearest neighbor matching \cite{Ho2007}) on age and sex to derive a control cohort of patients with any other admission besides I50*. This resulted in a total of 5,498 patients with 39,908 admissions. Within each admission there may be multiple measurements, in this analysis we take the mean of each measurement within an admission. The admission with the first heart failure diagnosis is selected and for the control cohort, the admission with the fewest missing values was selected. 

\subsection{Data Pre-processing}

We use Bidirectional Recurrent Imputation for Time Series (BRITS), which combines an imputation loss with the loss of a prediction or classification task \cite{Cao2018}, which we found to be most effective on previous cohorts. 
Each admission contains laboratory work and vital sign measurements. Vital signs and laboratory measures were selected if each measure was present in more than 60\% of cases. Features used in this work include systolic blood pressure, diastolic blood pressure, heart rate, oxygen saturation (SpO2), temperature, alanine aminotransferase (ALT), creatinine, c-reactive protein (CRP), platelets, potassium, sodium, urea and white blood cells.

Cases were then excluded if less than 60\% of these features were present for a particular admission. This resulted in a reduction in patients to 4,497 (2,298 heart failure and 2,199 control). A test set of 25\% was removed from the dataset in a stratified way and was held out of all training procedures. The remaining 75\% of the data was used in 5-fold cross validation in order to optimize the network parameters and ensure the model was not overfitting.

\section{Results and Discussion}

The accuracy of distinguishing heart failure and non-heart failure cases using the three approaches was determined through the area under the ROC curves as shown in Figure \ref{roc}. ROC curves are obtained by training a random forest on the PCA and DEC embedded spaces, whereas for DSEC the ROC curve is obtained from the semi-supervised classification step. DSEC obtains an area under the ROC curve of 0.84, considerably outperforming PCA (0.66) and DEC (0.73).

Figure \ref{hierarchical} shows a hierarchical clustering of the learnt DSEC embedding, where we iteratively combine the groups into larger subgroups in order to investigate whether different comorbidities exist within different spaces of the embedding. 

\begin{table}[h]
\caption{Enriched ICD10 codes between hierarchical splits in the clustering (log odds-ratio shown in brackets). Enrichment is performed between pairs of clusters, (for example 1 vs 2, 2.1 vs 2.2, and 1.1.1 vs 1.1.2). Sub-hierarchies of cluster 1 (control) are not shown.}
\label{enriched}
\vskip 0.15in
\begin{center}
\begin{small}
\begin{sc}
\centering
\begin{tabular}{cl}
\hline
Group & Enriched ICD-10 Codes \\
\hline
1 & E78.0 Pure hypercholesterolaemia (0.96) \\
\hline
2 & I50.0 Congestive heart failure (1.69) \\
 & E87.7 Fluid overload (1.39) \\
 & I50.9 Heart failure (1.32) \\
 & N18.9 Chronic renal failure (1.22) \\
 & I34.0 Mitral (valve) insufficiency (1.17) \\
\hline
2.1 & I42.0 Dilated cardiomyopathy (1.56) \\
 & N17.9 Acute renal failure (1.37) \\
 & N39.0 Urinary tract infection (1.34) \\
 & Z95.1 Aortocoronary bypass graft (1.10) \\
 & N18.9 Chronic renal failure (1.04) \\
\hline
2.2 & \\
\hline
2.1.1 & I50.1 Left ventricular failure (2.35) \\
\hline
2.1.2 & R18 Ascites (3.00) \\
& E87.5 Hyperkalaemia (2.13) \\
& I42.0 Dilated cardiomyopathy (1.55) \\
& I50.0 Congestive heart failure (1.16) \\
& N17.9 Acute renal failure (1.15) \\
\hline
2.1.1.1 & E87.5 Hyperkalaemia (3.26) \\
 & Z51.5 Palliative care (2.69) \\
\hline
2.1.1.2 & \\
\hline
2.2.1 & I27.2 Second. pulmonary hypertension (2.91) \\
& N18.9 Chronic renal failure (1.45) \\
& I48.9 Atrial fibrillation and flutter (1.36) \\
& E87.7 Fluid overload (1.09) \\
& Z92.1 Use of anticoagulants (0.95) \\
\hline
2.1.2 & \\
\hline
\end{tabular}
\end{sc}
\end{small}
\end{center}
\vskip -0.1in
\end{table}

Table \ref{enriched} shows cluster enrichment for the hierarchical splits. The first split in the hierarchical clustering is between heart failure (group 2) and controls (group 1). Subgroups of heart failure can be identified, including dilated cardiomyopathy, renal failure, and aortocoronary bypass grafts in a heart failure subgroup (group 2.1). Group 2.1 can be further split in to subgroups associated with left ventricular failure (2.1.1) and with ascites and hyperkalaemia (2.1.2). Group 2.2 can also be divided in to patients associated with secondary pulmonary hypertension, atrial fibrillation and flutter, and a history of anticoagulant use (2.2.1). This demonstrates that the method is capable of determining clinically relevant \cite{Maisel2003,Dickhout2011} subgroups from vital signs and laboratory measures. Further analysis of the subgroups and a comparison between PCA, DEC, and DSEC showing the superior performance of DSEC is shown in the supplementary materials.

While this is a powerful extension of the standard autoencoder embedding, which has been previously applied, current limitations are that we only consider a single admission per patient and the inability to deal with missing values, which are a common problem in EHRs.

We have shown our method outperforms other clustering algorithms to determine subgroups within heart failure patients. We aim to extend this approach to handle multiple admissions and to develop imputation free methods of embedding to further improve phenotyping of heart failure and other diseases from EHRs. This has the potential to allow for adjusted treatments of the different patient subgroups.

\section{Conclusions}

In this paper we demonstrate the application of DSEC to features derived from EHRs. We show our approaches can distinguish heart failure and non-heart failure cases based on laboratory measurements and vital signs. We illustrate that optimizing the embedding on known subgroups allows us to learn a more powerful representation and that subgroups within the heart failure cohort show enrichment of certain co-morbidities (ICD-10 codes). 

\section*{Acknowledgments}

This work uses data provided by patients and collected by the NHS as part of their care and support. We believe using patient data is vital to improve health and care for everyone and would, thus, like to thank all those involved for their contribution. The data were extracted, anonymised, and supplied by the Trust in accordance with internal information governance review, NHS Trust information governance approval, and General Data Protection Regulation (GDPR) procedures outlined under the Strategic Research Agreement (SRA) and relative Data Sharing Agreements (DSAs) signed by the Trust and Sensyne Health plc.

This research has been conducted using the Oxford University Hospitals NHS Foundation Trust Clinical Data Warehouse, which is supported by the NIHR Oxford Biomedical Research Centre and Oxford University Hospitals NHS Foundation Trust. Special thanks to Kerrie Woods, Kinga Varnai, Oliver Freeman, Hizni Salih, Zuzana Moysova, Professor Jim Davies and Steve Harris.

\bibliography{example_paper}

\begin{thebibliography}{31}
\providecommand{\natexlab}[1]{#1}
\providecommand{\url}[1]{\texttt{#1}}
\expandafter\ifx\csname urlstyle\endcsname\relax
  \providecommand{\doi}[1]{doi: #1}\else
  \providecommand{\doi}{doi: \begingroup \urlstyle{rm}\Url}\fi

\bibitem[Beaulieu-Jones \& Greene(2016)Beaulieu-Jones and
  Greene]{Beaulieu-Jones2016}
Beaulieu-Jones, B.~K. and Greene, C.~S.
\newblock {Semi-Supervised Learning of the Electronic Health Record for
  Phenotype Stratification}.
\newblock \emph{J Biomed Inform}, 64:\penalty0 168--178, 2016.

\bibitem[Beaulieu-Jones et~al.(2018)Beaulieu-Jones, Lavage, Snyder, Moore,
  Pendergrass, and Bauer]{Beaulieu-Jones2018}
Beaulieu-Jones, B.~K., Lavage, D.~R., Snyder, J.~W., Moore, J.~H., Pendergrass,
  S.~A., and Bauer, C.~R.
\newblock {Characterizing and Managing Missing Structured Data in Electronic
  Health Records: Data Analysis}.
\newblock \emph{JMIR Medical Informatics}, 6\penalty0 (1):\penalty0 e11, 2018.

\bibitem[Cao et~al.(2018)Cao, Zhou, Wang, Li, Li, and Li]{Cao2018}
Cao, W., Zhou, H., Wang, D., Li, Y., Li, J., and Li, L.
\newblock {BRITS: Bidirectional recurrent imputation for time series}.
\newblock \emph{Advances in Neural Information Processing Systems}, \penalty0
  (NeurIPS):\penalty0 6775--6785, 2018.

\bibitem[Choi et~al.(2016)Choi, Schuetz, Stewart, and Sun]{Choi2016}
Choi, E., Schuetz, A., Stewart, W.~F., and Sun, J.
\newblock Medical concept representation learning from electronic health
  records and its application on heart failure prediction.
\newblock 2016.
\newblock URL \url{http://arxiv.org/abs/1602.03686}.

\bibitem[Choi et~al.(2019)Choi, Xu, Li, Dusenberry, Flores, Xue, and
  Dai]{Choi2019}
Choi, E., Xu, Z., Li, Y., Dusenberry, M.~W., Flores, G., Xue, Y., and Dai,
  A.~M.
\newblock {Graph Convolutional Transformer: Learning the Graphical Structure of
  Electronic Health Records}.
\newblock pp.\  1--17, 2019.
\newblock URL \url{http://arxiv.org/abs/1906.04716}.

\bibitem[Denaxas et~al.(2018)Denaxas, Stenetorp, Riedel, Pikoula, Dobson, and
  Hemingway]{Denaxas2018}
Denaxas, S., Stenetorp, P., Riedel, S., Pikoula, M., Dobson, R., and Hemingway,
  H.
\newblock {Application of Clinical Concept Embeddings for Heart Failure
  Prediction in UK EHR data}.
\newblock 2018.
\newblock URL \url{http://arxiv.org/abs/1811.11005}.

\bibitem[Dickhout et~al.(2011)Dickhout, Carlisle, and Austin]{Dickhout2011}
Dickhout, J.~G., Carlisle, R.~E., and Austin, R.~C.
\newblock {Interrelationship between cardiac hypertrophy, heart failure, and
  chronic kidney disease: Endoplasmic reticulum stress as a mediator of
  pathogenesis}.
\newblock \emph{Circulation Research}, 108\penalty0 (5):\penalty0 629--642,
  2011.

\bibitem[Donders et~al.(2006)Donders, van~der Heijden, Stijnen, and
  Moons]{Donders2006}
Donders, A. R.~T., van~der Heijden, G.~J., Stijnen, T., and Moons, K.~G.
\newblock {Review: A gentle introduction to imputation of missing values}.
\newblock \emph{Journal of Clinical Epidemiology}, 59\penalty0 (10):\penalty0
  1087--1091, 2006.

\bibitem[Enguehard et~al.(2019)Enguehard, O'Halloran, and
  Gholipour]{Enguehard2019}
Enguehard, J., O'Halloran, P., and Gholipour, A.
\newblock {Semi-Supervised Learning With Deep Embedded Clustering for Image
  Classification and Segmentation}.
\newblock \emph{IEEE Access}, 7\penalty0 (1):\penalty0 11093--11104, 2019.

\bibitem[Fisher(1922)]{Fisher1922}
Fisher, R.~A.
\newblock {On the Interpretation of $\chi$ 2 from Contingency Tables, and the
  Calculation of P}.
\newblock \emph{Journal of the Royal Statistical Society}, 85\penalty0
  (1):\penalty0 87, 1922.

\bibitem[Han et~al.(2019)Han, Vedaldi, and Zisserman]{Han2019}
Han, K., Vedaldi, A., and Zisserman, A.
\newblock {Learning to Discover Novel Visual Categories via Deep Transfer
  Clustering}.
\newblock 2019.
\newblock URL \url{http://arxiv.org/abs/1908.09884}.

\bibitem[Ho et~al.(2007)Ho, Imai, King, and Stuart]{Ho2007}
Ho, D.~E., Imai, K., King, G., and Stuart, E.~A.
\newblock {Matching as nonparametric preprocessing for reducing model
  dependence in parametric causal inference}.
\newblock \emph{Political Analysis}, 15\penalty0 (3):\penalty0 199--236, 2007.

\bibitem[Inamdar \& Inamdar(2016)Inamdar and Inamdar]{Inamdar2016}
Inamdar, A. and Inamdar, A.
\newblock {Heart Failure: Diagnosis, Management and Utilization}.
\newblock \emph{Journal of Clinical Medicine}, 5\penalty0 (7):\penalty0 62,
  2016.

\bibitem[Kingma \& Ba(2015)Kingma and Ba]{Kingma2015}
Kingma, D.~P. and Ba, J.~L.
\newblock {Adam: A method for stochastic optimization}.
\newblock \emph{3rd International Conference on Learning Representations, ICLR
  2015 - Conference Track Proceedings}, pp.\  1--15, 2015.

\bibitem[Lasko et~al.(2013)Lasko, Denny, and Levy]{Lasko2013}
Lasko, T.~A., Denny, J.~C., and Levy, M.~A.
\newblock {Computational phenotype discovery using unsupervised feature
  learning over noisy, sparse, and irregular clinical data}.
\newblock \emph{PLoS One}, 8\penalty0 (6):\penalty0 e66341, 2013.

\bibitem[Li et~al.(2019)Li, Rao, Solares, Hassaine, Canoy, Zhu, Rahimi, and
  Salimi-Khorshidi]{Li2019}
Li, Y., Rao, S., Solares, J. R.~A., Hassaine, A., Canoy, D., Zhu, Y., Rahimi,
  K., and Salimi-Khorshidi, G.
\newblock {BEHRT: Transformer for Electronic Health Records}.
\newblock 2019.
\newblock URL \url{http://arxiv.org/abs/1907.09538}.

\bibitem[Liu et~al.(2019)Liu, Li, Hu, Shi, Wang, Tang, and Zhang]{Liu2019}
Liu, L., Li, H., Hu, Z., Shi, H., Wang, Z., Tang, J., and Zhang, M.
\newblock {Learning Hierarchical Representations of Electronic Health Records
  for Clinical Outcome Prediction}.
\newblock 2019.
\newblock URL \url{http://arxiv.org/abs/1903.08652}.

\bibitem[Maisel \& Stevenson(2003)Maisel and Stevenson]{Maisel2003}
Maisel, W.~H. and Stevenson, L.~W.
\newblock {Atrial fibrillation in heart failure: Epidemiology, pathophysiology,
  and rationale for therapy}.
\newblock \emph{American Journal of Cardiology}, 91\penalty0 (6):\penalty0
  2--8, 2003.

\bibitem[Miotto et~al.(2016)Miotto, Li, Kidd, and Dudley]{Miotto2016}
Miotto, R., Li, L., Kidd, B.~A., and Dudley, J.~T.
\newblock {Deep Patient: An Unsupervised Representation to Predict the Future
  of Patients from the Electronic Health Records}.
\newblock \emph{Scientific Reports}, 6:\penalty0 26094, 2016.

\bibitem[Nair \& Hinton(2010)Nair and Hinton]{Nair2010}
Nair, V. and Hinton, G.
\newblock {Rectified Linear Units Improve Restricted Boltzmann Machines}.
\newblock \emph{International Conference on Machine Learning}, 2010.

\bibitem[Ren et~al.(2019)Ren, Hu, Dai, Pan, Hoi, and Xu]{Ren2019}
Ren, Y., Hu, K., Dai, X., Pan, L., Hoi, S.~C., and Xu, Z.
\newblock {Semi-supervised deep embedded clustering}.
\newblock \emph{Neurocomputing}, 325:\penalty0 121--130, 2019.

\bibitem[Rokach \& Maimon(2005)Rokach and Maimon]{Rokach2005}
Rokach, L. and Maimon, O.
\newblock {Clustering Methods}.
\newblock In Maimon, O. and Rokach, L. (eds.), \emph{Data Mining and Knowledge
  Discovery Handbook}, pp.\  321--352. Springer US, Boston, MA, 2005.

\bibitem[Shickel et~al.(2018)Shickel, Tighe, Bihorac, and Rashidi]{Shickel2018}
Shickel, B., Tighe, P.~J., Bihorac, A., and Rashidi, P.
\newblock {Deep EHR: A Survey of Recent Advances in Deep Learning Techniques
  for Electronic Health Record (EHR) Analysis}.
\newblock \emph{IEEE J Biomed Health Inform}, 22\penalty0 (5):\penalty0
  1589--1604, 2018.

\bibitem[Vincent et~al.(2010)Vincent, Larochelle, Lajoie, Bengio, and
  Manzagol]{Vincent2010}
Vincent, P., Larochelle, H., Lajoie, I., Bengio, Y., and Manzagol, P.~A.
\newblock {Stacked denoising autoencoders: Learning Useful Representations in a
  Deep Network with a Local Denoising Criterion}.
\newblock \emph{Journal of Machine Learning Research}, 11:\penalty0 3371--3408,
  2010.

\bibitem[Ward \& Hook(1963)Ward and Hook]{Ward1963}
Ward, J.~H. and Hook, M.~E.
\newblock {Application of an Hierarchical Grouping Procedure to a Problem of
  Grouping Profiles}.
\newblock \emph{Educational and Psychological Measurement}, 23\penalty0
  (1):\penalty0 69--81, 1963.

\bibitem[Wei \& Eickhoff(2018)Wei and Eickhoff]{Wei2018}
Wei, X. and Eickhoff, C.
\newblock {Embedding Electronic Health Records for Clinical Information
  Retrieval}.
\newblock 2018.
\newblock URL \url{http://arxiv.org/abs/1811.05402}.

\bibitem[Wold et~al.(1987)Wold, Esbensen, and Geladi]{Wold1987}
Wold, S., Esbensen, K., and Geladi, P.
\newblock {Principal component analysis}.
\newblock \emph{Chemometrics and intelligent laboratory systems}, 2\penalty0
  (1-3):\penalty0 37--52, 1987.

\bibitem[Wong et~al.(2017)Wong, Wu, and Watkinson]{Wong2017}
Wong, D., Wu, N., and Watkinson, P.
\newblock {Quantitative metrics for evaluating the phased roll-out of clinical
  information systems}.
\newblock \emph{International Journal of Medical Informatics}, 105:\penalty0
  130--135, 2017.

\bibitem[Xie et~al.(2016)Xie, Girshick, and Farhadi]{Xie2015}
Xie, J., Girshick, R., and Farhadi, A.
\newblock {Unsupervised deep embedding for clustering analysis}.
\newblock \emph{33rd International Conference on Machine Learning, ICML 2016},
  1:\penalty0 740--749, 2016.
\newblock URL \url{http://arxiv.org/abs/1511.06335}.

\bibitem[Zhang et~al.(2018)Zhang, Kowsari, Harrison, Lobo, and
  Barnes]{Zhang2018}
Zhang, J., Kowsari, K., Harrison, J.~H., Lobo, J.~M., and Barnes, L.~E.
\newblock {Patient2Vec: A Personalized Interpretable Deep Representation of the
  Longitudinal Electronic Health Record}.
\newblock \emph{IEEE Access}, 6:\penalty0 65333--65346, 2018.

\bibitem[Zhu et~al.(2016)Zhu, Yin, Qian, Cheng, Wei, and Wang]{Zhu2016}
Zhu, Z., Yin, C., Qian, B., Cheng, Y., Wei, J., and Wang, F.
\newblock {Measuring patient similarities via a deep architecture with medical
  concept embedding}.
\newblock In \emph{2016 IEEE 16th International Conference on Data Mining
  (ICDM)}, pp.\  749--758. IEEE, 2016.

\end{thebibliography}
\bibliographystyle{icml2020}

\end{document}